# Short-term Load Forecasting at Different Aggregation Levels with Predictability Analysis

Yayu Peng[1,2], Yishen Wang[1], Xiao Lu[3], Haifeng Li[3], Di Shi[1], Zhiwei Wang[1], Jie Li[4]
[1]GEIRI North America, San Jose, CA, United States
[2]University of Nebraska, Lincoln, NE, United States
[3]State Grid Jiangsu Electric Power Company Ltd., Nanjing, Jiangsu, China
[4]State Grid US Representative Office, New York City, NY, United States
Email: yishen.wang@geirina.net

*Abstract*— **Short-term load forecasting (STLF) is essential for the reliable and economic operation of power systems. Though many STLF methods were proposed over the past decades, most of them focused on loads at high aggregation levels only. Thus, low-aggregation load forecast still requires further research and development. Compared with the substation or city level loads, individual loads are typically more volatile and much more challenging to forecast. To further address this issue, this paper first discusses the characteristics of small-and-medium enterprise (SME) and residential loads at different aggregation levels and quantifies their predictability with approximate entropy. Various STLF techniques, from the conventional linear regression to state-of-the-art deep learning, are implemented for a detailed comparative analysis to verify the forecasting performances as well as the predictability using an Irish smart meter dataset. In addition, the paper also investigates how using data processing improves individual-level residential load forecasting with low predictability. Effectiveness of the discussed method is validated with numerical results.**

*Index Terms*—**Deep learning, machine learning, smart meter, short-term load forecasting, load aggregation level, predictability**

## I. INTRODUCTION

Electricity load forecasting is of great importance for the reliable and economic power systems operation and planning as well as efficient market management [1][2]. This paper specifically focuses on short-term load forecasting (STLF) technologies to support operators in the energy management system applications and other decision-making processes [3]-[5].

Numerous researchers have proposed various forecasting models for STLF at transmission levels. Since 1990s, machine learning based load forecasting techniques have gained much more popularity in the field. Multiple linear regression methods utilized historic load data, weather data, day type and other context information to gain accurate forecasts and they are widely adopted in today's utilities and ISOs [1] . In addition, methods including artificial neural networks (ANN) with multi-layer perceptron (MLP), support vector machine (SVM) [3], random forest (RF) [6], gradient boosted regression tree (GBRT) [7], gaussian processes [7] have also achieved significant improvement compared to classic time series models in various load forecasting competitions [8]. Recently, deep learning has become a popular research topic and its applications in load forecasting started around 2014. Among them, deep neural network (DNN) and recurrent neural networks (RNN) provide the most popular architectures. In [9], DNN and traditional machine learning approaches are applied and the results indicate that certain DNN architectures achieve better accuracy than traditional methods. Similarly, authors in [10] explore various combinations of activation functions and shows significant improvement using the ELU function. Compared with conventional feedforward neural networks, RNN has the particular advantage to cope with historical data through a feedback connection. As an extension of RNN, long-short-term memory (LSTM) has been introduced in load forecasting area in the past few years [11]-[13].

Traditionally, STLF is conducted to forecast high-voltage level loads such as the one in a substation or the whole metropolitan area. On the other hand, forecasting of load at lower voltage level is rarely conducted due to the lack of high-quality data. Fortunately, the massive deployment of smart meters creates the opportunities to forecast and analyze customer consumption at a much lower level with much higher granularity [14]. By aggregating individual smart meter readings, loads at different aggregation levels can be obtained and analyzed. Because of the volatile nature of individual customer, load forecasting at low level is much more challenging, and therefore, the forecasting methods are still in the development stage [13]. The characteristics of individual load and aggregated loads with the Aggregation Error Curve (AEC) are discussed in [15].

Considering such load characteristics difference, it is important to discuss the predictability before applying forecasting algorithms. For low aggregation level residential loads, the load pattern is always dominated by residents' behaviors and most of their behaviors are highly stochastic which results in the low predictability. Due to such facts, all forecasting methods have relatively large forecasting errors on

This work is funded by SGCC Science and Technology Program under contract no. SGSDYT00FCJS1700676.

this type of load, no matter how advanced the forecasting methods are and how delicately their hyperparameters are tuned. Rather than using alternative advanced learning algorithms, one feasible way to reduce the forecasting error for this type of load is to process the data to improve its predictability.

This paper first discusses the characteristics of residential and SME loads at different aggregation levels in detail, from individual meter load to highly aggregated load. In addition, this paper quantifies the predictability of load at different aggregation levels and investigates the effectiveness and performances of state-of-the-art load forecasting techniques, ranging from simple linear regression to complex LSTM. To further improve the forecasting with low predictability time series, data differencing is adopted to illustrate the importance of conducting predictability analysis and applying data processing techniques.

The rest of the paper is organized as follows. Section II discusses the load forecasting problem formulation and evaluation metrics. Section III analyzes the characteristics of individual residential and SME load as well as the predictability at different aggregation. Implementation of forecasting techniques and results are discussed in Section IV. Finally, conclusions are drawn in Section V.

## II. MACHINE LEARNING-BASED LOAD FORECASTING TECHNIQUES

This section first introduces the load forecasting problem, and then forecasting and evaluation techniques used in this work.

### A. Load forecasting

Depending on the input features, there are three common approaches for load forecasting. The first one is physical-based forecasting, using temperature, humidity, time, wind speed, etc. The second one is statistical-based forecasting, using historical load data. The third one is the hybrid of the previous two. Theoretically, the third one should be able to provide better results as more information is incorporated. However, the second approach is still commonly used because of the ease in implementation. Such convenience is very important, especially when the contextual factors are unavailable or costly to retrieve.

The methods evaluated in this paper only include features from historical load data due to data availability. Without losing generality, other contextual information such as weather, temperature, illumination, etc., can be easily incorporated as additional features to further improve the performance of STLF with the same framework [16], [17].

The load is denoted by a sequence of data points $x(t)$, where $t$ is the time index indicating when the load is measured. Load forecasting can then be formulated as using values at $x(t)$, $x(t-1)$,…, $x(t-N)$ to predicted the values $M$ steps ahead $\hat{x}(t+M)$ as shown in (1), where $f(x)$ can be any machine learning technique.

$$\hat{x}(t+M) = f\big(x(t), x(t-1), \ldots, x(t-N)\big) \quad (1)$$

The evaluation framework of STLF at different aggregation levels is shown in Fig. 1. First, data cleaning is conducted on the original smart meter data to remove incomplete and abnormal readings. Then, data are classified according to load types, namely, residential load and SME load. Next, loads are aggregated at different levels to generate the training, validation and testing set. Finally, a one-hour-ahead and one-day ahead forecasting are conducted using different forecasting techniques.

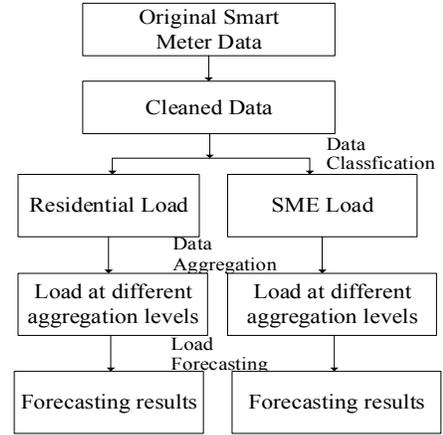

Fig. 1. Framework of the STLF at different load aggregation level

### B. Machine learning-based load forecasting techniques

In this paper, several state-of-the-art forecasting algorithms are evaluated and compared, including linear regression, gradient boosted regression tree (GBRT), support vector regression (SVR), multi-layer perceptron (MLP) and long-short-term memory (LSTM). Interested readers can refer to the references for more details [1]-[8],[18]-[22].

### C. Evaluation metric

The most widely used metrics in load forecasting literature are root mean square error (RMSE), mean absolute error (MAE) and mean absolute percentage error (MAPE). The RMSE is scale-dependent and is not suitable for comparing forecasting results at different aggregation levels. The MAPE is scale-independent, but the drawback is the normalization term in the denominator may be close to zero for individual smart meter readings (for residential load, it is common to have load values close to zero when the resident is not at home), leading to very high MAPE values. Therefore, this paper adopts normalized mean absolute error (NMAE) shown in (2) as evaluation metric for forecasting performance [18].

$$NMAE(x, \hat{x}) = \frac{MAE(x,\hat{x})}{\|x\|_1} = \frac{\sum_{t=1}^{n}|x(t)-\hat{x}(t)|}{\sum_{t=1}^{n}|x(t)|} \quad (2)$$

## III. CHARACTERISTICS OF LOAD AT DIFFERENT AGGREGATION LEVELS

This section discusses the characteristics of residential and SME loads as well as their predictability and the effect of load aggregation.

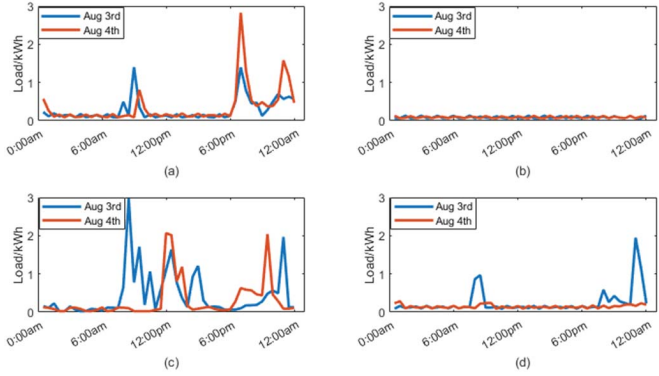

Fig. 2 Typical individual residential load

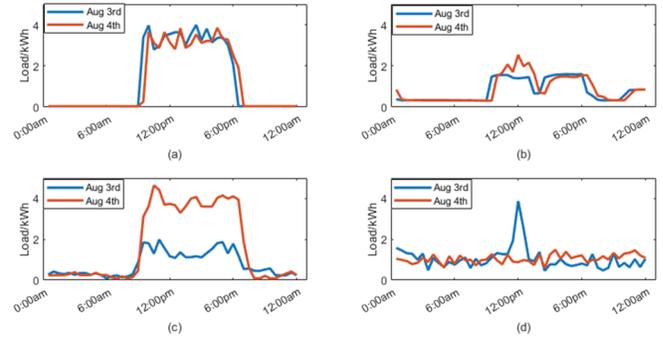

Fig. 3 Typical individual SME load

## A. Data description

Data used in this paper is from Commission for Energy Regulation (CER) in Ireland. They consists of smart meter readings for over 5,000 Irish homes and small business between 2009 and 2010 [19]. After data cleaning, a total of 1700 residential and 250 SME loads are used in this paper. The smart meter data are collected every 30 minutes; therefore, one-hour and one-day ahead load forecasting correspond to 2-step and 48-step ahead prediction. The customer types in this dataset are classified as 1) Residential, 2) SME, and 3) others. For data analysis purpose, this paper focuses on residential and SME load.

## B. Residential load

Forecasting residential load at individual level is challenging because of the high randomness involved. The energy consumption at each household heavily depends on the lifestyle of its resident(s). Fig. 2 (a)-(d) presents selected typical 2-day (August 3$^{rd}$ and 4$^{th}$, 2009) residential load patterns for meters whose IDs are 1058, 1065, 1067, 1178 respectively. Fig. 2 (a) shows high peaks around 9:00-10:00am, 6:30-7:30pm, 10:30-11:30pm which may be related to cooking breakfast, dinner preparation, and taking bath. Even though the electricity consumption pattern of the second day is relatively similar to the first day, there still exists some differences. However, the residential load can be very different for different households or even the same households at different times. Fig. 2 (b) shows a low load pattern, it is very likely that the house is vacant and the electricity is consumed by the refrigerator. In Fig. 2 (c), the load pattern on the second day is different from the first day. This is common because uncertainty is dominant at individual loads. Behaviors of residents can be changing. Fig. 2 (d) shows two peaks in the morning and evening for the first day and the load is very low and does not have any peak on the second day. It is likely that the resident is at home the first day and not at home on the second day. In general, individual residential load depends heavily on the resident lifestyle and forecasting such pattern accurately at smart meter levels is very challenging without having other data.

## C. SME load

Individual SME load behaves more regularly than individual residential load because most of SME have their regular working hours. Fig. 3 (a)-(d) show typical SME loads for smart meters whose IDs are 5132, 3088, 1050, 7195 respectively. It is obvious that the load patterns of the next day are similar to the previous day even though load patterns look different for different meters. This periodicity makes the forecasting of individual SME load easier than residential load. However, there still exists users whose electricity consumption is random as shown in Fig. 3 (d). Moreover, from the fact that most of the SMEs work from Monday to Friday only, the weekday and weekend load can be forecasted separately with different models and parameters. This paper targets forecasting of weekday load for SMEs. In addition, according to the magnitudes of the SME, these loads are more likely to be small shops not large business.

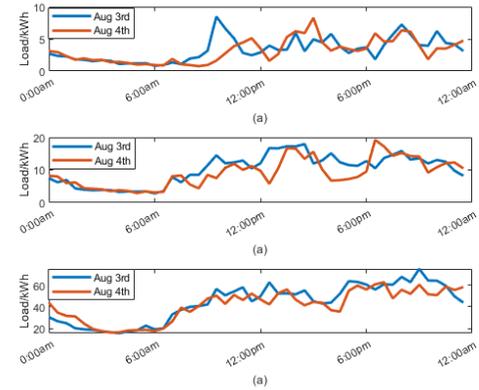

Fig. 4 Residential load at different aggregation level (a)5 (b)20 (c)100 smart meter readings

## D. Predictability and effects of load aggregation

Loads at different aggregation levels can be obtained by aggregating different numbers of smart meter readings together. As the randomness is alleviated and smoothed after aggregation, load forecasting in this case becomes easier. Fig. 4 shows the aggregation of residential loads with 5, 20, and 100 smart meter readings, respectively. In Fig. 4(a), the randomness still dominates the load pattern even though the load is more periodic and smoother than individual loads. As the aggregation level increases, the next-day load shape is more similar to the previous day, indicating the increase in periodicity makes the load forecasting easier. Moreover, the aggregation operation serves as a low-pass filter, removing the high frequency impulses corresponding to random events in the load curve. Fig. 5 shows the aggregated SME loads with 5, 20, and 100 smart meter readings, respectively. Their periodicity appears at a much lower aggregation level than residential load; this is because the individual SME loads are already more periodic than individual residential loads. In general, the effect of aggregation is reducing the randomness of load and can make the load more predictable.

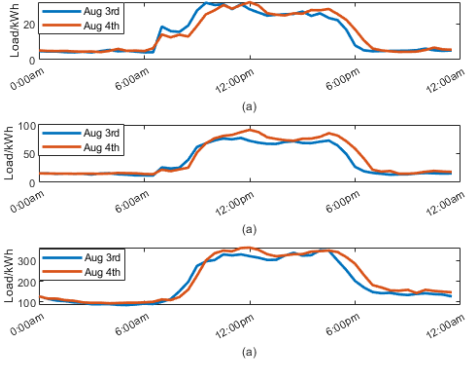

Fig. 5 SME load at different aggregation level (a) 5 (b) 20 (c) 100 smart meter readings

Except for the law of large numbers, such effect can be explained with the complexity and predictability of time series. The predictability of time series spreads a wide range. On the low end of this range are time series that exhibit perfect predictive structure, i.e., constant or periodic series. On the high end of this predictability range, series are called fully complex, where underlying generating process transmits no information at all from the past to the future. White noise falls into this class. In fully complex series, knowledge of the past gives no insight of the future, regardless of the model. Therefore, predictability analysis provides insights on forecasting performances, and it is important to discuss and analyze load predictability at different aggregation levels for residential and SME loads [20].

In this paper, the predictability is measured by approximate entropy (*ApEn*) as suggested in [20]. A small *ApEn* implies that the sequence is regular and more predictable, whereas a high *ApEn* indicates a more random pattern. Fig. 6 (a) presents the *ApEn* for residential load at different aggregation levels. The *ApEn* is very high at low aggregation levels and decreases drastically as the aggregation level increases and then stays stable after the aggregation level reaches 1200. This illustrates that the residential load is very irregular at individual levels and becomes more predictable as aggregation level increases. Since *ApEn* stays stable at high aggregation levels, it can be inferred that forecasting performance can be hardly improved further without introducing new information. Fig. 6 (b) shows the *ApEn* for SME loads at different aggregation levels. The initial low value of *ApEn* and its decreasing trend suggest that the SME loads are much more predictable even at individual level.

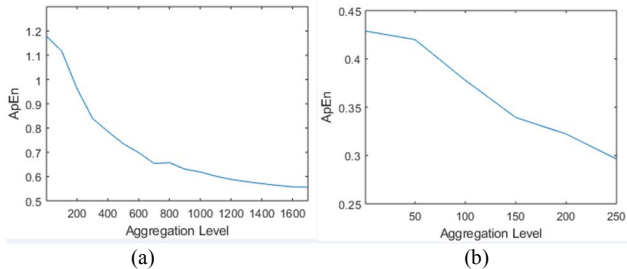

Fig. 6 ApEn of (a) residential (b) SME load at different aggregation levels

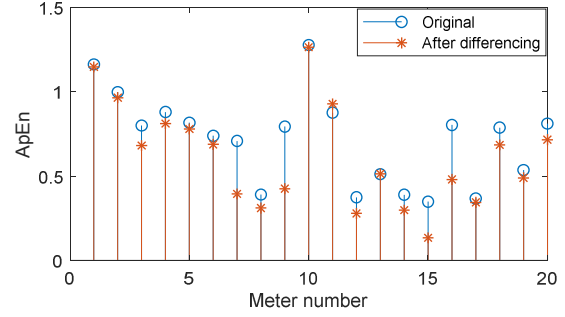

Fig. 7 ApEn of original data and data after processing

### E. *Data processing for individual-level residential load*

As shown in Fig. 2(a)-(d), the individual residential load is highly stochastic. The peak load appears at different times and the load pattern of the second day could be totally different from the previous day. As shown in Fig. 6(a), such *ApEn* is the highest among all cases. Therefore, predicting this type of load is naturally challenging. One common approach is to use more sophisticated forecasting models, but it is more likely to fit the noise in the data and has bad generalization capability even though its training error can be very low. Instead of using such approach, this paper advocates conducting data processing before applying forecasting methods. The effectiveness of the data processing technique can be quantified by the proposed predictability metric *ApEn*.

The selection of data processing techniques heavily depends on the data itself. There does not exist one method that works for all datasets. In this paper, the differencing is adopted as an example to show the power on data processing for improving predictability. More advanced forecasting algorithms can be used for further analysis of the processed data.

Fig. 7. shows the ApEn for 20-meter readings as an example. For most meter readings (18/20), the *ApEn* of the data after processing is lower than the *ApEn* of original data, which means the data processing indeed increases the predictability of data even with simple differencing. As discussed earlier, there are various developed data processing methods. There does not exist a method that applies to all datasets, and the researchers need to tailor the processing method according to the data to achieve the maximum predictability improvements.

## IV. STLF ON LOADS AT DIFFERENT AGGREGATION LEVELS

A total of 100 days of historical load data are used, among 70% is used for training, 20% is for validation, and the remaining 10% is for testing. The input features are the historical load of the past 10 days. One-hour-ahead and one-day-ahead forecasting are both evaluated. The techniques introduced in section II are implemented on loads at different aggregation levels and their performance are compared. Different techniques have multiple hyperparameters that need tuning to achieve good performance. For GBRT, the hyperparameters are learning rate, number of estimators and maximum depths. For SVR, the radial basis function (RBF) kernel is used. The hyperparameters are the penalty of the error term and the tolerance range for the predicted values from the actual ones. For neural networks with structure MLP, the number of hidden layers, number of hidden neurons in each layer, number of epochs and batch sizes all need tuning. For neural networks with structure LSTM, the number of LSTM blocks, batch sizes and number epochs are tuned. To find the satisfying hyperparameter set, grid search algorithm is used for the tuning. These algorithms are all implemented with Python under Windows.

The aggregation error curves (AEC) as shown in Fig. 8 and Fig. 9 show how the errors decrease with the increase of aggregation levels and how each forecasting technique behaves at different levels. With the help of AEC, appropriate forecasting techniques can be evaluated and selected.

### A. Forecasting results for residential load

Fig. 8 (a) and (b) show the result of one-hour-ahead and one-day-ahead forecasting for residential loads at different aggregation levels. It is obvious that the forecasting errors decrease with the increase of aggregation levels. Moreover, the errors of one-day-ahead forecasting is much higher than one-hour-ahead forecasting when the aggregation level is low. This can be explained by the randomness of individual human behaviors. The historical load cannot provide the full spectrum of information. For example, if the residents will go vacation tomorrow, this will cause dramatic load changes which cannot be inferred from historical load data without new information. No matter what techniques are used, the forecasting errors on low aggregation level residential loads are always large. This is due to the low predictability of the data series itself as the *ApEn* shown in Fig. 6. Therefore, simply by fine tuning different forecasting techniques can hardly further increase forecasting accuracy. One feasible approach to increase the accuracy is to incorporate contextual information such as the temperature, humidity, travel plans or even the daily life pattern of residents.

The error starts decreasing with the increase of aggregation level for both one-hour and one-day ahead forecasting. This is due to the increase of predictability. The performance of each forecasting technique is slightly different. For residential dataset, linear regression, SVR, MLP have similar performance over all aggregation levels and are better than GBRT. This is because GBRT cannot output values larger than its training data and there exists a slight increasing trend in the datasets.

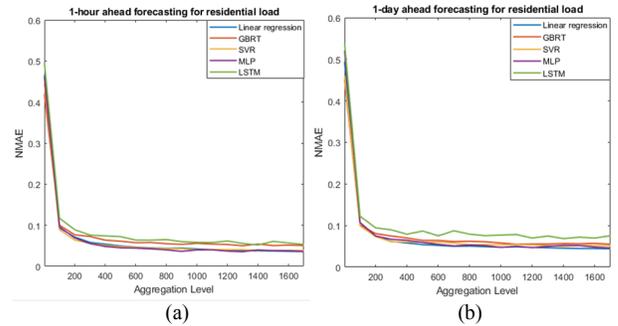

Fig. 8 AEC of residential load for (a) 1-hour (b) 1-day ahead forecasting

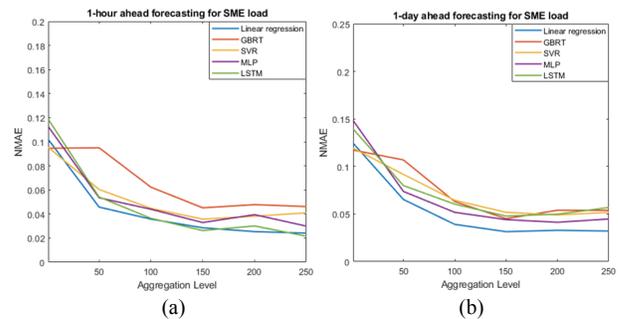

Fig. 9 AEC of SME load for (a) 1-hour (b) 1-day ahead forecasting

### B. Forecasting results for SME load

Similarly, Fig. 9 (a) and (b) show results of one-hour-ahead and one-day-ahead forecasting for SME loads at different aggregation levels. The initial error of SME load is lower than the initial error of residential load. This can be interpreted from the computed *ApEn* shown in Fig. 6 that the individual SME load is more regular than individual residential load. As expected, the error is also decreasing with the increase of aggregation levels. In addition, GBRT is less capable than the Linear regression, MLP and SVR.

### C. Forecasting results for original data and processed data

As discussed in the previous section, the data differencing is an effective method to reduce *ApEn* for some residential smart meter readings. The readings of the 20 meters are used to compare the forecasting error using the original data and processed data and the results are shown in Fig. 10. It can be concluded from the two figures that for most meters, the forecasting error is reduced using the processed data instead of the original data. However, there still exists some meter readings that forecasting error is not reduced. The reason is discussed earlier that there does not exist one data processing technique that is effective for all datasets. The selection of data processing technique should take the characteristics of datasets into consideration and it is dataset dependent.

## V. CONCLUSIONS & DISCUSSIONS

This paper firstly discusses the characteristics of residential and SME loads at individual meter level and the effects of load aggregation. In addition, the predictability of loads at different aggregation levels are discussed. Various machine learning techniques have also been applied to the hour-ahead and day-ahead STLF for these two types of loads. To evaluate techniques at each load aggregation level, hyperparameters are carefully analyzed and tuned. With the collected smart meter data, case studies verify that not only the load types, but also the load aggregation levels significantly affect the forecasting performance. At low aggregation levels, due to the low predictability, all algorithms have relatively large errors, especially for residential loads. With the increase of aggregation levels, the predictability increases as well as the performance of all algorithms. The effect of aggregation is more notable for residential loads than SME loads.

Selection of load forecasting techniques highly depends on the data itself, and there is no single technique that outperforms other techniques in all scenarios. It is recommended to conduct predictability analysis on data before directly applying any forecasting techniques, especially for load at low aggregation levels.

For the individual-level residential load forecasting, simply tuning forecasting parameters or adopting more sophisticated algorithms can barely improve forecasting performance when the time series predictability does not improve. One approach is to conduct data processing to improve the predictability of the data. This paper demonstrates that even the simple differencing can be an effective processing technique to improve the predictability of data. Other more advanced processing techniques, such as wavelet decomposition and empirical mode decomposition, can also be applied to increase predictability. However, load forecasting is a data-centric problem, and the selection of data processing technique should be based on the characteristics of the data. There does not exist a technique that works for all.

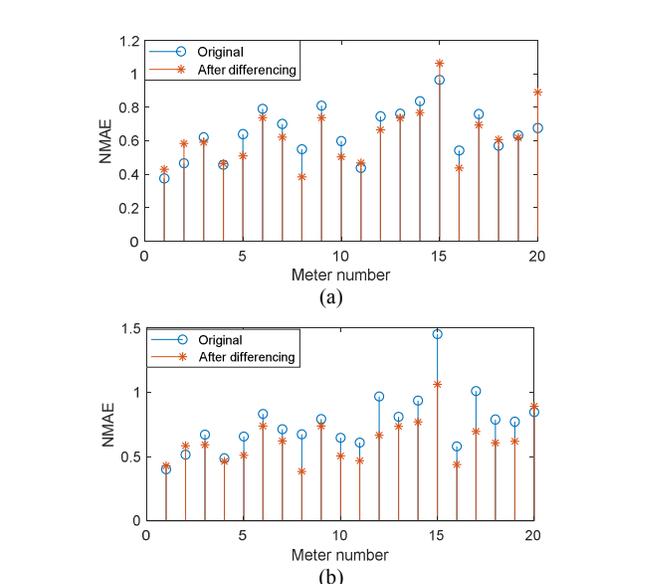

Fig. 10 NMAE of individual level residential load forecasting using: a) linear regression, b) SVR.